\PassOptionsToPackage{unicode}{hyperref}
\PassOptionsToPackage{hyphens}{url}
\PassOptionsToPackage{dvipsnames,svgnames,x11names}{xcolor}
\documentclass[
  12pt]{article}

\usepackage{amsmath,amssymb}
\usepackage{iftex}
\ifPDFTeX
  \usepackage[T1]{fontenc}
  \usepackage[utf8]{inputenc}
  \usepackage{textcomp} % provide euro and other symbols
\else % if luatex or xetex
  \usepackage{unicode-math}
  \defaultfontfeatures{Scale=MatchLowercase}
  \defaultfontfeatures[\rmfamily]{Ligatures=TeX,Scale=1}
\fi
\usepackage{lmodern}
\ifPDFTeX\else  
\fi
\IfFileExists{upquote.sty}{\usepackage{upquote}}{}
\IfFileExists{microtype.sty}{% use microtype if available
  \usepackage[]{microtype}
  \UseMicrotypeSet[protrusion]{basicmath} % disable protrusion for tt fonts
}{}
\makeatletter
\@ifundefined{KOMAClassName}{% if non-KOMA class
  \IfFileExists{parskip.sty}{%
    \usepackage{parskip}
  }{% else
    \setlength{\parindent}{0pt}
    \setlength{\parskip}{6pt plus 2pt minus 1pt}}
}{% if KOMA class
  \KOMAoptions{parskip=half}}
\makeatother
\usepackage{xcolor}
\makeatletter
\ifx\paragraph\undefined\else
  \let\oldparagraph\paragraph
  \renewcommand{\paragraph}{
    \@ifstar
      \xxxParagraphStar
      \xxxParagraphNoStar
  }
  \newcommand{\xxxParagraphStar}[1]{\oldparagraph*{#1}\mbox{}}
  \newcommand{\xxxParagraphNoStar}[1]{\oldparagraph{#1}\mbox{}}
\fi
\ifx\subparagraph\undefined\else
  \let\oldsubparagraph\subparagraph
  \renewcommand{\subparagraph}{
    \@ifstar
      \xxxSubParagraphStar
      \xxxSubParagraphNoStar
  }
  \newcommand{\xxxSubParagraphStar}[1]{\oldsubparagraph*{#1}\mbox{}}
  \newcommand{\xxxSubParagraphNoStar}[1]{\oldsubparagraph{#1}\mbox{}}
\fi
\makeatother

\usepackage{longtable,booktabs,array}
\usepackage{calc} % for calculating minipage widths
\usepackage{etoolbox}
\makeatletter
\patchcmd\longtable{\par}{\if@noskipsec\mbox{}\fi\par}{}{}
\makeatother
\IfFileExists{footnotehyper.sty}{\usepackage{footnotehyper}}{\usepackage{footnote}}
\makesavenoteenv{longtable}
\usepackage{graphicx}
\makeatletter
\def\maxwidth{\ifdim\Gin@nat@width>\linewidth\linewidth\else\Gin@nat@width\fi}
\def\maxheight{\ifdim\Gin@nat@height>\textheight\textheight\else\Gin@nat@height\fi}
\makeatother
\setkeys{Gin}{width=\maxwidth,height=\maxheight,keepaspectratio}
\makeatletter
\def\fps@figure{htbp}
\makeatother

\makeatletter
\@ifpackageloaded{caption}{}{\usepackage{caption}}
\AtBeginDocument{%
\ifdefined\contentsname
  \renewcommand*\contentsname{Table of contents}
\else
  \newcommand\contentsname{Table of contents}
\fi
\ifdefined\listfigurename
  \renewcommand*\listfigurename{List of Figures}
\else
  \newcommand\listfigurename{List of Figures}
\fi
\ifdefined\listtablename
  \renewcommand*\listtablename{List of Tables}
\else
  \newcommand\listtablename{List of Tables}
\fi
\ifdefined\figurename
  \renewcommand*\figurename{Figure}
\else
  \newcommand\figurename{Figure}
\fi
\ifdefined\tablename
  \renewcommand*\tablename{Table}
\else
  \newcommand\tablename{Table}
\fi
}
\@ifpackageloaded{float}{}{\usepackage{float}}
\floatstyle{ruled}
\@ifundefined{c@chapter}{\newfloat{codelisting}{h}{lop}}{\newfloat{codelisting}{h}{lop}[chapter]}
\floatname{codelisting}{Listing}

\makeatother
\makeatletter
\@ifpackageloaded{caption}{}{\usepackage{caption}}
\@ifpackageloaded{subcaption}{}{\usepackage{subcaption}}
\makeatother

\ifLuaTeX
  \usepackage{selnolig}  % disable illegal ligatures
\fi
\usepackage[]{natbib}
\usepackage{bookmark}

\IfFileExists{xurl.sty}{\usepackage{xurl}}{} % add URL line breaks if available
\hypersetup{
  pdftitle={Title},
  pdfauthor={Author 1; Author 2},
  pdfkeywords={3 to 6 keywords, that do not appear in the title},
  colorlinks=true,
  linkcolor={blue},
  filecolor={Maroon},
  citecolor={Blue},
  urlcolor={Blue},
  pdfcreator={LaTeX via pandoc}}

\newcommand{\anon}{1}

\begin{document}

\def\spacingset#1{\renewcommand{\baselinestretch}%
{#1}\small\normalsize} \spacingset{1}

%%%%%%%%%%%%%%%%%%%%%%%%%%%%%%%%%%%%%%%%%%%%%%%%%%%%%%%%%%%%%%%%%%%%%%%%%%%%%%

\if1\anon
{
  \title{\bf "Rebuilding" Statistics in the Age of AI: A Town Hall Discussion on Culture, Infrastructure, and Training}

  \author{
    David L.\ Donoho\textsuperscript{1}, 
    Jian Kang\textsuperscript{2}, 
    Xihong Lin\textsuperscript{3}, 
    Bhramar Mukherjee\textsuperscript{4}, \\
    Dan Nettleton\textsuperscript{5},
    Rebecca Nugent\textsuperscript{6}, 
    Abel Rodriguez\textsuperscript{7}, 
    Eric P.\ Xing\textsuperscript{8},\\ 
    Tian Zheng\textsuperscript{9},  and 
    Hongtu Zhu\textsuperscript{10}\thanks{  
    \textsuperscript{1}Department of Statistics, Stanford University; 
    \textsuperscript{2}Department of Biostatistics, University of Michigan, Ann Arbor; 
    \textsuperscript{3}Harvard T.H. Chan School of Public Health; Department of Statistics, Harvard University; Broad Institute; 
    \textsuperscript{4}Yale School of Public Health; Department of Statistics and Data Science, Yale University; 
    \textsuperscript{5}Department of Statistics, Iowa State University; 
    \textsuperscript{6}Department of Statistics and Data Science, Carnegie Mellon University; 
    \textsuperscript{7}Baskin School of Engineering, University of California, Santa Cruz; 
    \textsuperscript{8}Mohamed bin Zayed University of Artificial Intelligence; School of Computer Science, Carnegie Mellon University; 
    \textsuperscript{9}Department of Statistics, Columbia University; 
    \textsuperscript{10}Department of Biostatistics,  University of North Carolina at Chapel Hill.\\
    \textit{The authors are listed alphabetically.}
    }
          \date{}  
  }
  \maketitle
} \fi

\if0\anon
{
  \bigskip
  \bigskip
  \bigskip
 \title{\bf "Rebuilding" Statistics in the Age of AI: A Town Hall Discussion on Culture, Infrastructure, and Training}
  \author{} 
  \date{} 
   \maketitle
 
  \medskip
} \fi

\bigskip
\begin{abstract}
This  article presents the full, original record of the 2024 Joint Statistical Meetings (JSM) town hall, “Statistics in the Age of AI,” which convened leading statisticians to discuss how the field is evolving in response to advances in artificial intelligence, foundation models, large-scale empirical modeling, and data-intensive infrastructures. The town hall was structured around open panel discussion and extensive audience Q\&A, with the aim of eliciting candid, experience-driven perspectives rather than formal presentations or prepared statements. This document preserves the extended exchanges among panelists and audience members, with minimal editorial intervention, and organizes the conversation around five recurring questions concerning disciplinary culture and practices, data curation and “data work,” engagement with modern empirical modeling, training for large-scale AI applications, and partnerships with key AI stakeholders. By providing an archival record of this discussion, the preprint aims to support transparency, community reflection, and ongoing dialogue about the evolving role of statistics in the data- and AI-centric future.

%This perspective paper distills and elaborates on key themes from the 2024 JSM town hall, “Statistics in the Age of AI,” providing a more complete account of the discussion in which panelists examined how the field should evolve as foundation models, large-scale empirical modeling, and data-intensive infrastructures reshape science and society. Rather than providing a transcript or survey, we organize the expanded discussion around five recurring questions: (i) disciplinary culture and incentives; (ii) the central role of data curation and annotation; (iii) constructive engagement with modern empirical modeling; (iv) training for large-scale AI applications; and (v) partnerships with key AI stakeholders. We argue that sustaining the discipline’s influence will require statisticians to broaden both their technical toolkit and institutional footprint—embedding data fairness, cloud-scale computation, and data engineering into routine practice; strengthening evaluation, uncertainty quantification, and interpretability for high-stakes AI systems; and prioritizing end-to-end, system-level problem solving over narrowly scoped contributions. We conclude with pragmatic recommendations for reshaping training, incentives, and collaboration so that statisticians help lead—not merely support—the data-centric future.

\end{abstract}

\noindent%
\vfill

\newpage
\spacingset{1.8} % DON'T change the spacing!

\section{Introduction}

The rapid expansion of artificial intelligence (AI) and large-scale data technologies is transforming scientific research, industry, and society at an unprecedented pace \citep{moor2023foundation,goodfellow2016deep,
vaswani2017attention,PearsonLedfordHutsonVanNoorden2025MostCited,NSF2025ExpandingNAIRRDataSystems,NIHODSS2025ArtificialIntelligence,NSF2026AIInstitutes,WhiteHouse2025GenesisMissionFactSheet}. Foundation models, large-scale empirical modeling, and data-intensive infrastructures increasingly shape decision-making across domains ranging from health and environmental science to economics and public policy. These developments present both opportunities and pressures for the statistical community \citep{he2025statistics,LinCaiDonohoEtAl2025StatisticsAI,Breiman2001TwoCultures,IMS2025PresidentialAddressStatisticsCrossroadsAI,ASA2025NSFRFIAIRD}: while statistics remains central to uncertainty quantification, causal reasoning, and principled evaluation, many traditional practices, institutional structures, and training models were not designed for the scale, speed, and engineering complexity characteristic of modern AI systems.

Against this backdrop, we organized a town hall session entitled \emph{``Statistics in the Age of AI: A Town Hall with an Expert Panel''} at the 2024 Joint Statistical Meetings (JSM). The session convened a distinguished panel of statisticians—Drs. David Donoho, Xihong Lin, Bhramar Mukherjee, Dan Nettleton, Rebecca Nugent, Abel Rodriguez, and Eric Xing—and was moderated by Drs. Tian Zheng and Hongtu Zhu. Rather than formal presentations, the town hall was intentionally structured around open discussion and extensive audience Q\&A, with the goal of eliciting candid, experience-driven perspectives on how statistics, statistics and biostatistics departments, and statisticians themselves are being empowered, challenged, and reshaped by advances in AI.

This article provides the full, original record of that town hall conversation. It preserves the extended exchanges among panelists, follow-up questions, and audience interactions, with minimal editorial intervention beyond light organization and clarification. Our aim is not to offer a retrospective synthesis or normative framework, but to document the discussion as it unfolded, capturing both points of consensus and areas of tension. In doing so, we seek to create an archival resource for the statistical community—one that reflects how leaders in the field are reasoning, debating, and grappling with the implications of AI in real time.

\begin{figure}[htbp]
    \centering
    \includegraphics[width=\textwidth]{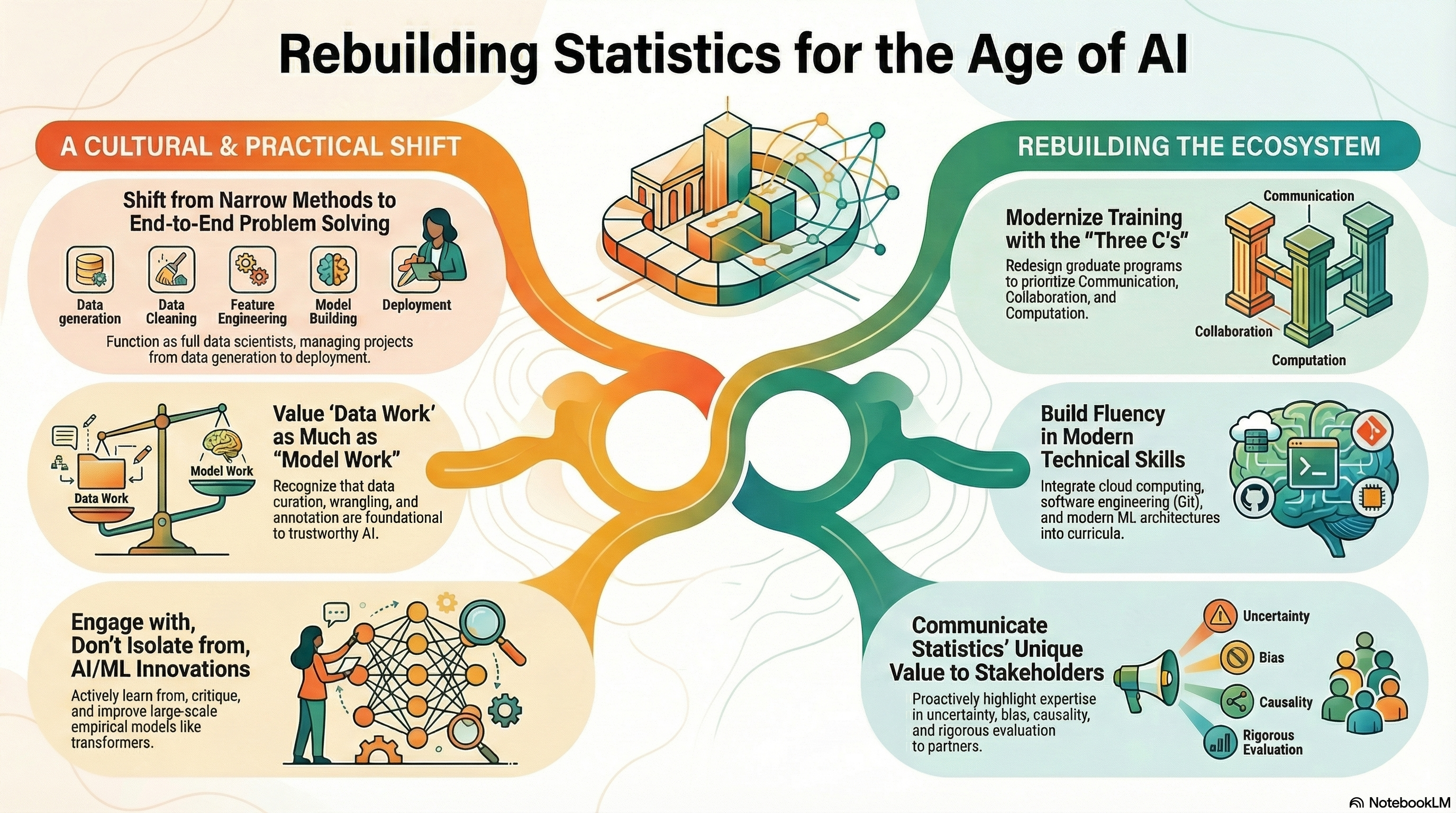}
    \caption{Key themes emerging from the 2024 JSM town hall on ``Statistics in the Age of AI.'' The figure highlights a cultural and practical shift toward end-to-end problem solving; the centrality of data work alongside model development; active engagement with modern AI/ML innovations; modernization of training around communication, collaboration, and computation; fluency in modern technical skills; and clearer communication of statistics’ unique value—such as uncertainty quantification, bias assessment, and causal reasoning—to stakeholders. This figure was created by using Google NotebookLM. }
    \label{fig:stats_ai_townhall}
\end{figure}

Figure~\ref{fig:stats_ai_townhall} provides a visual summary of several cross-cutting themes that emerged repeatedly during the town hall discussion. These include a shift from narrowly scoped methodological contributions toward end-to-end, system-level problem solving; explicit recognition of data curation, annotation, and engineering as foundational components of trustworthy AI; deeper engagement with modern empirical and algorithmic modeling; modernization of training around communication, collaboration, and computation; and clearer articulation of statistics’ distinctive strengths—such as uncertainty quantification, bias assessment, causal reasoning, and rigorous evaluation—to a broad range of stakeholders. While the town hall conversation ranged widely, these themes collectively frame the questions and perspectives documented in the sections that follow.

\begin{figure}[htbp]
    \centering
    \includegraphics[width=\textwidth]{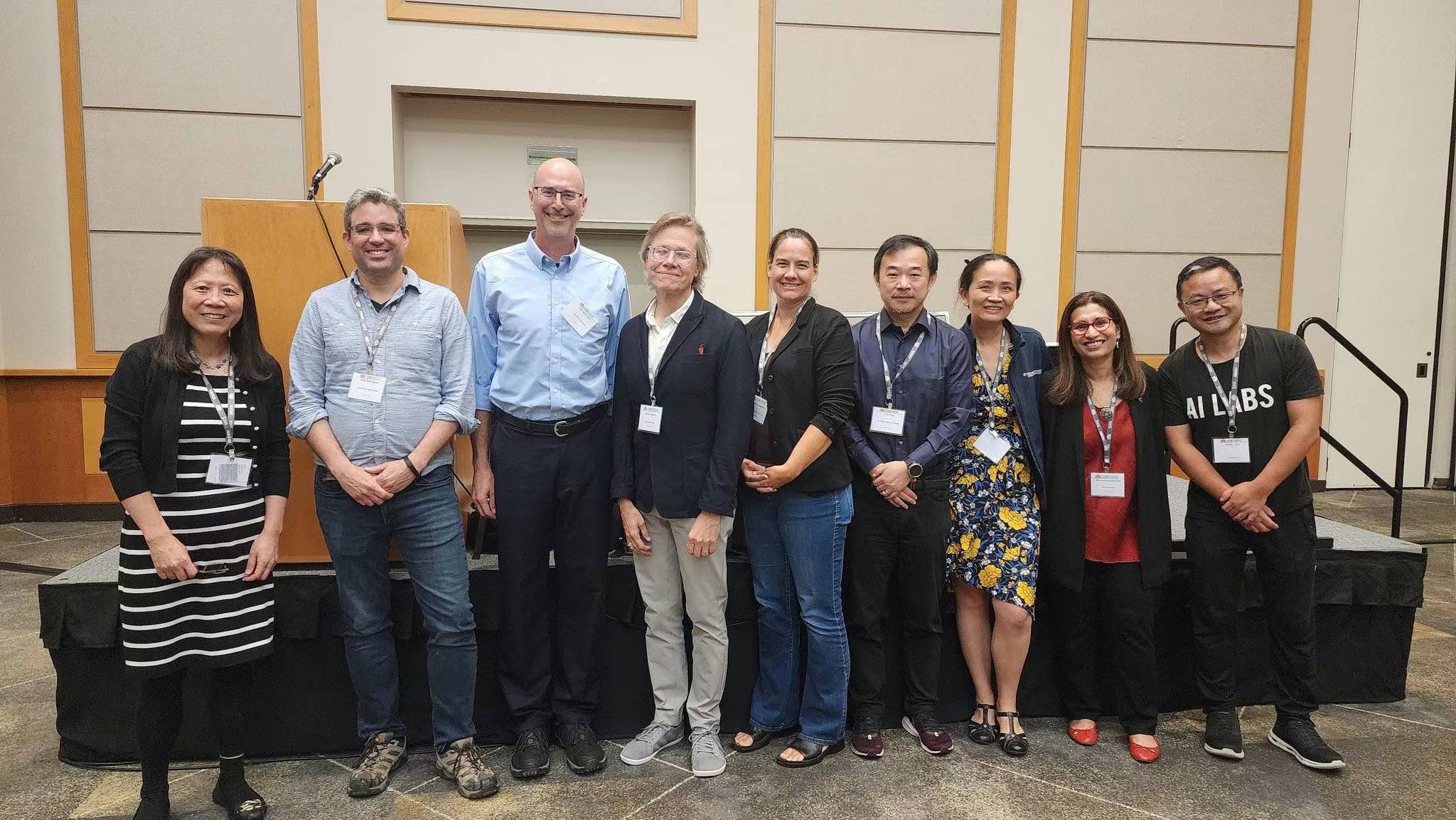}\\[2mm]
    \includegraphics[width=0.49\textwidth]{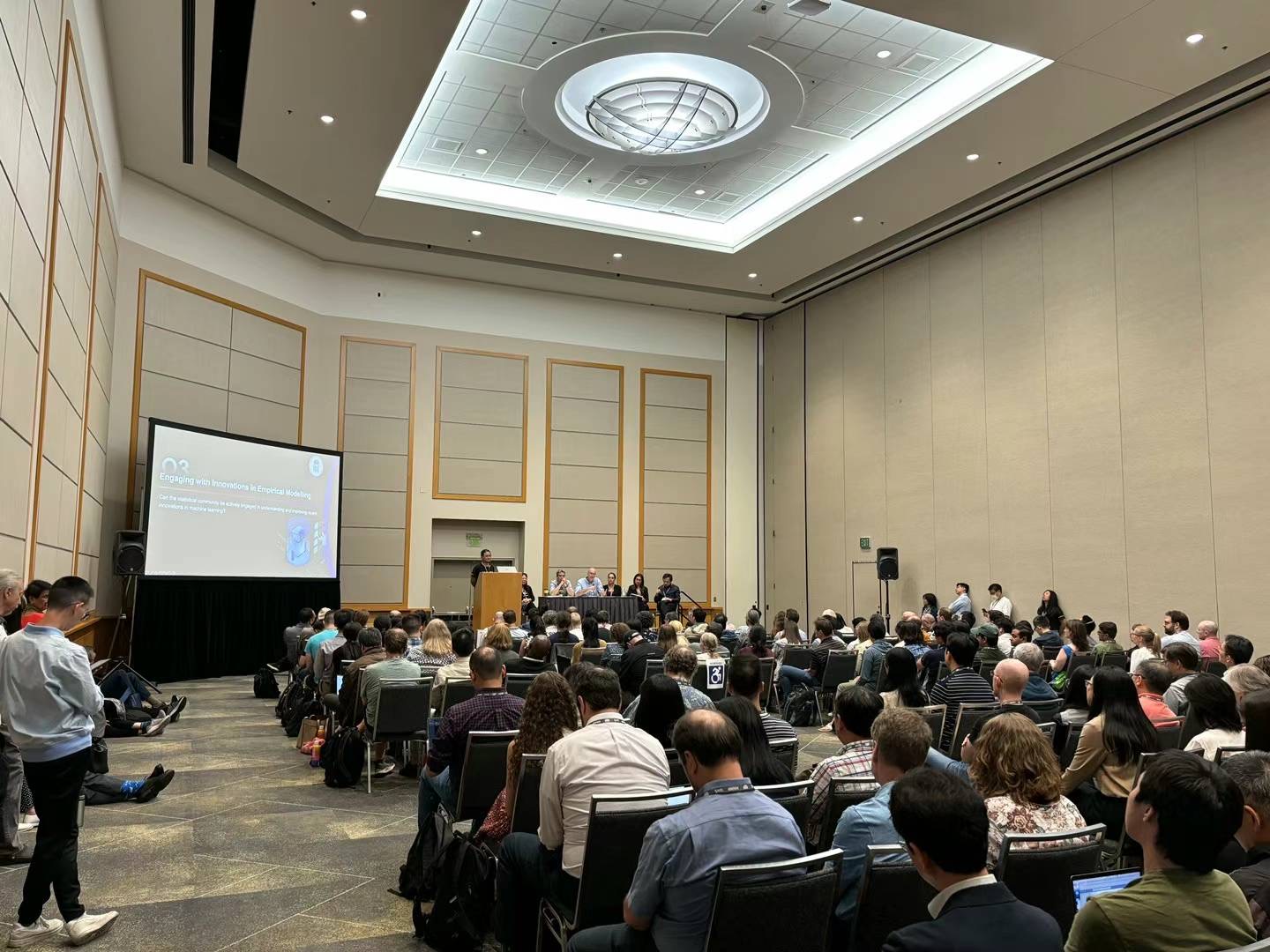}\hfill
    \includegraphics[width=0.49\textwidth]{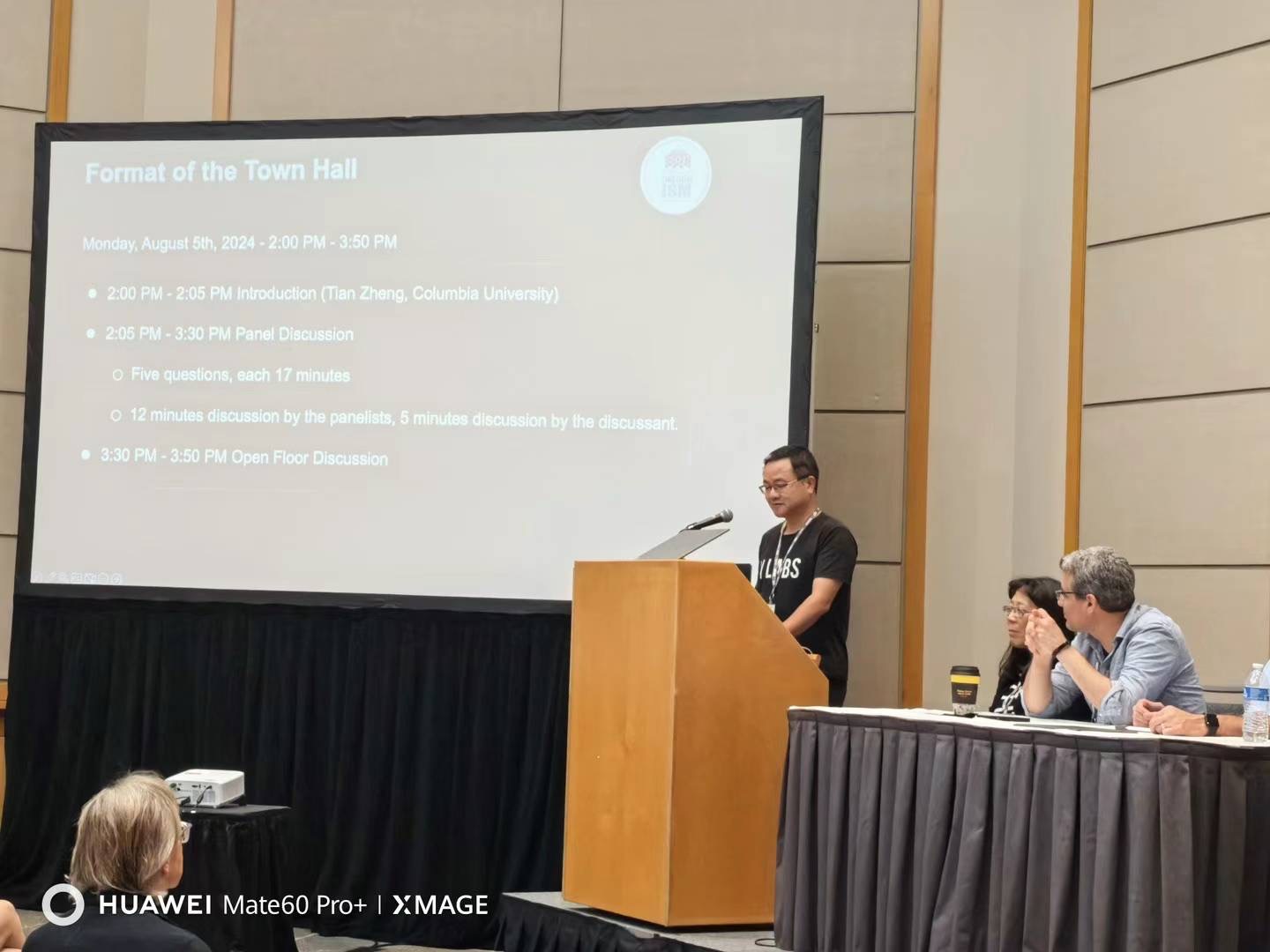}
    \caption{
    %Scenes from the 2024 Joint Statistical Meetings town hall
    %``Statistics in the Age of AI.'' \textbf{Top:} A panelist addressing the audience.
    %\textbf{Bottom left:} A wide view of the session highlighting audience size and engagement.
     %\textbf{Bottom right:} The town hall format emphasizing open discussion and audience Q\&A.
    %These images illustrate the interactive, community-driven nature of the event.
    Highlights from the 2024 Joint Statistical Meetings (JSM) town hall, ``Statistics in the Age of AI.'' \textbf{Top}: Panelists and organizers representing diverse perspectives across statistics and AI. \textbf{Bottom left}: A wide view of the session demonstrating the high level of attendance and community interest. \textbf{Bottom right}: Professor Hongtu Zhu introducing format of the Town Hall.
    }
    \label{fig:townhall_context}
\end{figure}

Figure~\ref{fig:townhall_context} provides visual context for the town hall setting,
format, and audience, underscoring the interactive and community-oriented nature of
the discussion documented in this article.

For readability, the conversation is organized around five interrelated questions that structured the town hall and repeatedly surfaced during the discussion:
\begin{itemize}
    \item \textbf{Question (i) (Culture and Practices of Statistics):} How should the culture, practices, and infrastructure of statistics evolve to engage effectively with large-scale data processing and analysis? What challenges must the field address to remain relevant and impactful?

    \item \textbf{Question (ii) (Data Curation and ``Data Work''):} How can statisticians deepen their engagement with data curation, wrangling, and annotation, and more fully recognize the central role of this ``data work'' in method development and AI systems?

    \item \textbf{Question (iii) (Engaging with Innovations in Empirical Modeling):} How can the statistical community actively understand, critique, and improve recent advances in machine learning and empirical modeling, rather than treating them as external or purely engineering-driven developments?

    \item \textbf{Question (iv) (Next-Generation Talent for Large-Scale Applications):} What strategies are most effective for training the next generation of statisticians to work in large-scale data and AI settings? What programs or initiatives have demonstrated success in cultivating these skills?

    \item \textbf{Question (v) (Engaging AI Stakeholders):} How can statisticians engage constructively with key AI stakeholders—including academia, industry, funding agencies, and policymakers—to promote statistical principles, attract talent and resources, and build sustainable collaborative ecosystems?
\end{itemize}

By making this discussion publicly available in its original form, we hope to support transparency, community reflection, and ongoing dialogue about the evolving role of statistics in the age of AI. We view this document as a complement to subsequent perspective and synthesis articles, and as a durable reference for researchers, educators, and leaders seeking to understand how the field is navigating this period of rapid change.

\section{Culture and Practices of Statistics}

The culture and everyday practices of statistics must evolve to remain impactful in the age of AI. Rather than viewing AI as an external threat or a separate discipline, the panel emphasized the need for statisticians to rethink their roles, workflows, and technical repertoire in the context of large-scale, engineering-heavy data ecosystems \citep{NSF2025ExpandingNAIRRDataSystems,NIHODSS2025ArtificialIntelligence,NSF2026AIInstitutes,WhiteHouse2025GenesisMissionFactSheet}.

\paragraph*{Xihong Lin:} I will start by sharing some of my thoughts about our culture and the practice and infrastructure of statistics. There are two takeaways I want to make. First, I think it would be helpful to have a mindset change—specifically, to think about how to build an end-to-end data science ecosystem. From my perspective, this involves three pillars. One is diverse and fair data. Second is building statistical machine learning and AI methods, tools, and infrastructure. And the third is making data analysis results interpretable. This will require significant change for ourselves, including how to become not just statisticians, but full data scientists with expertise in statistics as well as domain science. It also requires improving communication not only with domain scientists, but increasingly with computer scientists. Ultimately, the goal is for people from different fields to work together and make an impact in science by solving big, real-world problems.

Next, I will talk about data fairness. I feel that data fairness outweighs data size. Data size itself is not the issue. For big data, bias is much more important than variance. In particular, we need to make greater efforts to increase diversity in study samples.

I will use polygenic risk scores as an example to illustrate data fairness. Polygenic risk scores have been widely used in the past two years to predict disease risk using genotype profiles. However, over the past 20 years, 86\% of GWAS studies have been conducted in European populations. If one looks at the performance of these prediction models, using Europeans as a reference, and then applies the same models to Asian populations, the accuracy drops to about 50\%. When applied to African populations, the accuracy is only about 25\%. This highlights the importance of both data fairness and algorithmic fairness, and helps explain why there is so much interest in transfer learning right now.

The second takeaway is that, to truly build an end-to-end data science ecosystem, we need to engage more deeply in data science engineering. An end-to-end ecosystem requires automation throughout the process, including data allocation, analytic pipeline construction, and deployment in the cloud. I will talk more about data allocation in Question (ii). The computer science and machine learning communities are very open: they see statistics as useful and are willing to invest time in learning statistical concepts. The question is whether statisticians should also spend more time learning data science engineering.

Regarding the cloud environment, for whole-genome sequencing data and modern data science ecosystems, almost everything is now cloud-based. In the past, data sharing largely involved copying data—for example, UK Biobank data could be copied across institutions. This model no longer works because whole-genome sequencing data are too large. The current model instead brings researchers to the data rather than bringing data to the researchers. Multiple cloud platforms, such as Google Cloud and others, are now in use, and researchers need to become familiar with workflow languages and workspace-based environments. NIH and UK Biobank have strongly promoted the development of data clouds and data ecosystems. For example, ALS\_FTD data are available through AnVIL, and TOPMed data are available through BioData Catalyst. These resources are all cloud-based. UK Biobank data can no longer be downloaded and are now accessed through secure platforms. As a result, statisticians need to become familiar with these cloud environments and make their methods and tools available on such platforms.

\paragraph*{Dan Nettleton:} Statistics is the science of learning from data. I think that is the definition we should embrace for our field, if we have not already. Learning from both small and large datasets is important. Our discipline has contributed much over the last century in theory and methods for learning from datasets that fit nicely on our laptops. Such contributions will continue to be important for analyzing experimental data generated every day by scientists working both in and out of academia. We need to continue to be experts in traditional topics, such as experimental design and generalized linear mixed-effects modeling, to support scientific progress.

However, to address the data challenges of today, we need to broaden the reach of our discipline. To accomplish this broadening, the intersection—the knowledge that every statistician must know—may need to become smaller so that the union—the collective knowledge of all statistians—can grow larger. It seems unlikely that we will be allotted more time to train each statistics PhD student, so it is important not to insist that all of our students be clones of a single archetype when determining their training. I would like to draw an (admittedly imperfect) analogy to random forests. The genius of Leo Breiman’s random forests lies in the insistence that the trees in a forest be different from one another. Aside from working with slightly different training data due to bootstrap sampling, each tree is forced to use the data differently by considering different subsets of explanatory variables at each split. If the trees are all the same, a random forest is just a single tree and far less successful as a predictor. Analogously, if we view members of our statistical community as trees, our collective efforts as a community—a forest—will be more impactful when individuals share some common foundations while also possessing diverse and complementary skills and knowledge.

\paragraph*{Abel Rodriguez:} Thank you for the invitation to participate in the panel; I am very excited to be here. In addition to thinking about how we need to change, it is also important to reflect on what has made us successful and what distinguishes us from other disciplines. Our focus on the quantification and management of uncertainty is one clear area of strength for statisticians. The fact that we care not only about point prediction, but also about interval prediction and estimation, distinguishes both our training and the way we approach problems. Sadly, we sometimes struggle to communicate to non-statisticians why this is important. For example, I often think about the “reproducibility crisis” as an “uncertainty management” crisis. The fact that most statisticians think carefully about the process of data collection and potential sources of data bias (such as missing values, non-representative samples, causality, and the need to model correlations) is another major strength. Finally, statisticians are trained to care deeply about the application or domain science in which methods are applied or developed.

Returning to the original question, statisticians still lack training in what I would call “intermediate computer literacy.” Even relatively simple skills—such as the use of distributed version control systems (e.g., Git), basic software engineering principles, and APIs for accessing online data—are not regularly taught. Perhaps learning how to use large language models as part of the statistical pipeline will soon join this list. Unlike 15 or even 10 years ago, many of our undergraduate and graduate students now go on to industry positions, and the lack of these skills can be a disadvantage for them. In my opinion, it also hurts academic statisticians, as it limits our ability to disseminate new methodology and encourage broader adoption. Finally, while we are often very good at working with case studies—examining a single dataset in depth—we are not always as strong at thinking about systems in which methods are deployed repeatedly without expert supervision.

\paragraph*{Eric Xing:} Let me begin by rethinking the role and purpose of statistics as a scientific discipline.

Traditionally, statistics has served as a discipline that connects real data (experience) to knowledge through summarization, abstraction, and theorization, helping people understand and use data for a wide range of tasks. These tasks largely fall into two categories: (1) prediction (e.g., outcomes of certain events) and (2) discovery (of mechanisms or “causes”). It is important to point out that the often stand-alone goal of “understanding” in statistics, or in any science, is by itself subjective and insubstantial unless it is ultimately grounded in value through prediction or discovery.

Predictive and actionable understanding should therefore be the ultimate value we seek to extract from data, rather than equations, mathematical formulations, or theories that are attainable only for a small portion of the data universe. Yet many statisticians—and perhaps even the discipline as a whole—find aspiration and intellectual satisfaction primarily in such abstractions. This focus has shaped the culture we see today, reinforced by a highly championed identity centered on rigor, elegance, and logical structure, but with relatively diminished emphasis on utility—possibly influenced by pure mathematics. Much modern statistical work relies heavily on assumptions about data and scenarios in order to enable mathematical reasoning or proofs. However, in real-world environments and applications, few of these simplifying assumptions actually hold. This raises a fundamental question: do we discard data, abandon tasks, or avoid challenging problems in order to preserve traditional statistical practice? Or do we let go of a fixed identity as “statisticians” and become open to studying new data, new tools, and new problems that may contradict established norms?

Large language models (LLMs) epitomize this challenge \citep{vaswani2017attention,chang2024survey,worldmodel_simulative_reasoning_2025,llm_from_scratch_2025,song2024aido}. They use enormous datasets—effectively all text produced by civilization—while making few assumptions about data distributions. They are extremely high-dimensional, with trillions of parameters, defying conventional notions of the curse of dimensionality. They are trained using a very general self-supervised learning task, next-token prediction, without customization for specific downstream applications such as standardized tests. They make little use of prior linguistic knowledge from classical natural language processing, such as syntactic or semantic rules or topic models. Instead, they rely on a highly sophisticated yet generic and compute-friendly architecture—the transformer—built on mechanisms such as attention and layered vector projections, rather than complex statistical models like hidden Markov models or probabilistic graphical models that attempt to encode temporal or causal logic. They are optimized for computational efficiency rather than mathematical rigor, using low-precision representations, dropping pathological data points (e.g., those causing numerical errors), and leveraging distributed asynchronous computing across massive GPU clusters. They are also subject to extensive post hoc engineering, including fine-tuning, retrieval-augmented generation, and user-based alignment—practices that run counter to traditional statistical expectations of consistency in data usage throughout training.

Despite all this, LLMs achieve remarkable performance on well-defined tasks such as question answering and standardized tests across nearly all domains, including law, medicine, and even mathematics, surpassing previous computational systems and, in some cases, human performance. It is not inconceivable that similar technologies—transformer-based or transformer-inspired models—will produce comparable disruptions in domains where classical statistics currently plays an active role, such as medicine, biology, finance, and environmental science. Statisticians should therefore turn their attention to these emerging AI technologies, leveraging their strengths in mathematics and data analysis, but not necessarily in the ways traditionally emphasized by mainstream statistical journals such as \emph{The Annals of Statistics} or \emph{JASA}. Instead, they should engage more deeply with areas such as data engineering, software engineering, computational efficiency, environmental impact, and user behavior, bringing a statistical perspective to these issues and making a distinctive contribution.

This shift requires statisticians to:
\begin{enumerate}
\item \textbf{Think big}—in terms of problems (multi-task or general-purpose), data (billions or trillions of observations), and computation (yottaFLOPs, $10^{24}$, across tens of thousands of GPUs). We must rethink the balance between curiosity-driven research and goal-oriented or objective-driven research. What should the objective be: optimizing for the next paper in \emph{The Annals of Statistics}, achieving high predictive performance even without full explainability (as in the case of LLMs), or discovering new scientific laws? How should we balance explainability, actionability, and operationalizability?

\item \textbf{Reconsider fundamentals in scientific philosophy}, such as what constitutes “understanding” and “knowledge.” These questions were debated by Kantians and Humeans during the Enlightenment and must be revisited in the era of modern generative AI, which approaches problems very differently from first-principles rationalism. A prime example is AlphaFold’s use of transformer models for protein structure prediction, which does not rely on molecular dynamics equations grounded in atomic physics and chemistry. As Hume argued, all knowledge is based on experience. AI systems demonstrate that knowledge need not take the form of laws or mathematical equations alone. Algorithms, protocols, and operational procedures can also represent valid forms of knowledge or understanding. A winemaker who reliably produces high-quality wine using a generationally inherited process may possess knowledge no less legitimate than that of a chemist or mathematician building classifiers for wine.

\item \textbf{Avoid over-fixation on abstract terminology} such as intelligence, causality, optimality, uncertainty quantification, bounds, and convergence. These terms can create an illusion of understanding and cause researchers to stop too early, rather than probing more deeply into how models actually work on real data and real problems.

\item \textbf{Focus on original, real problems}, rather than always reducing them into abstractions. Instead of assuming limited data or computation, consider scenarios with near-infinite data, near-infinite compute, and near-infinite retrievable memory—conditions under which conceptually simple approaches, such as $k$-nearest neighbors or asymptotic nonparametric models, may become feasible. These should not be dismissed as “engineering details” or “messy artifacts.” The final mile of problem solving—production, deployment, and real-world action—must be walked by someone, and statisticians should be among those who do so.

\item \textbf{Step outside the echo chamber}. Statisticians need to engage more with data generators, engineers, and users to better understand where data come from, how they are processed, and how outputs are consumed. This broader engagement provides context and enables meaningful external validation of statistical work.
\end{enumerate}

From a technical standpoint, modern AI thrives on new learning paradigms such as multi-task learning, federated learning, and reinforcement learning; new loss functions such as contrastive loss, reward-based objectives, and utility-driven criteria; and architectures that implement complex procedures rather than models built primarily on assumptions and priors. Instead of emphasizing distribution fitting and likelihood optimization, modern AI draws heavily on differential geometry (manifolds, subspaces), information theory (coding, compression, reproducing kernel Hilbert spaces), linear algebra (spectral analysis, dynamical systems, tensor algebra), optimization, and variational calculus. Visualization tools—such as spectral methods and phylogenetic representations—can also play an important role in building human intuition and insight. These are tools that a modern statistical curriculum should increasingly incorporate.

In summary, modern statisticians should focus on entire projects and objectives rather than narrow slices, set aside rigid disciplinary identities, and use whatever methods are necessary—simple or complex—to solve real problems. We should be open to alternative ways of establishing credibility beyond abstract mathematical proofs. There are many ways to build trust. If I pour a glass of water safely 100 times, do I need to present fluid dynamics equations each time? Is that not sufficient evidence? Returning to the winemaking example, does one need to be a chemist to demonstrate expertise? Rigor means matching experience reliably and repeatedly, with or without equations. A procedure or algorithm that consistently produces dependable results represents a valid form of understanding. Rigor in the sense of methodology is not the north star—the problem being solved is.

\section{Data Curation and ``Data Work''}

Data curation, wrangling, and annotation are not merely support tasks but foundational work that shapes bias, uncertainty, interpretability, and the trustworthiness of AI systems. The panel urged statisticians to engage directly with how data are generated and processed at scale, and to realign training, infrastructure, and incentives so high-quality data resources and reproducible pipelines are valued alongside new methods.

\paragraph*{Bhramar Mukherjee:} A persistent complaint in data science is that “everyone wants to do the model work, nobody wants to do the data work.” And yet, data work is essential and increasingly dominant. Data annotation, data cleaning, and data curation play a crucial role in the development of many AI-related products, such as drones, self-driving cars, and large language models like ChatGPT. Overemphasis on method development can lead to both practical and theoretical disconnects between the mindset of young statisticians and the challenges they ultimately face. How can we avoid this?

\noindent\textbf{Summary:} Doing the dirty work is essential.
\begin{itemize}
\item We need to dismantle some aspects of our traditional thinking and reward systems to reimagine what we consider to be a strong statistical contribution. For example, statisticians played a key role in bringing together the OHDSI (Observational Health Data Science and Informatics) consortium, an effort that has paid dividends for years across the global EHR research community.

\item Another important point is the need for statisticians to be at the helm of data collection efforts and the leadership of large scientific studies. The University of Michigan biobanking initiative was started and led by a biostatistician. Designing and building resources such as NIH All of Us or the UK Biobank is no small feat and represents an enormous contribution to scientific research.

\item Finally, we need to think seriously about building staff scientist teams that are motivated and skilled in implementing cutting-edge data analysis for large-scale problems. Traditional academic settings focus primarily on students and faculty, but we also need strong teams of full-time data scientists within academia and clear career paths for them. This is essential for advancing large projects. Every department should have support for developing software packages and APIs, as well as providing AI/ML consultation for students.
\end{itemize}

\paragraph*{Xihong Lin:} I agree with many of the points that Bhramar made. I will share a few stories to illustrate the importance of data annotation. The first example I want to discuss is ImageNet. Many of you have probably heard of ImageNet; it contains over 10 million hand-annotated images and truly jump-started the modern deep learning revolution in AI. Before ImageNet, deep learning methods often underperformed compared to existing approaches because available datasets were simply not large enough. Building ImageNet was an enormous effort, led by Fei-Fei Li, who is currently co-director of Human-centered Artificial Intelligence at Stanford. She started this project in 2006 as an assistant professor. At that career stage, deciding to undertake the hand annotation of millions of images required tremendous courage. At the time, most researchers were focused on developing AI methods and tools, but she recognized that large, carefully annotated datasets were essential to move AI methods to the next level.

I will also share my own experience with data annotation efforts. To explain why this is important, consider the timeline of whole-genome sequencing. Over the past decade, sequencing technologies have changed dramatically. Multiple large-scale whole-genome sequencing programs have been launched, and I have been heavily involved in NHGRI-supported efforts. To date, we have sequenced about 350{,}000 individuals, and the Trans-Omics for Precision Medicine (TOPMed) program includes roughly another 200{,}000 individuals. Overall, more than half a million people have been sequenced. At the same time, multiple large biobanks have emerged. The most well-known is the UK Biobank, which integrates GWAS and whole-genome sequencing data with electronic medical records. As I mentioned earlier, UK Biobank data can no longer be downloaded; everything is now cloud-based. Another major effort is the Million Veteran Program (MVP), which has enrolled one million participants. A third is the NIH \emph{All of Us} Research Program, originally launched as the Precision Medicine Initiative under the Obama administration. Importantly, all of these large sequencing programs have made substantial efforts to increase diversity in their study populations.

With these massive datasets, we realized a few years ago that we needed to make a serious investment in functional annotation. To give you a sense of scale, we built a functional annotation database covering approximately 9 billion variants. Across the genome, there are about 3 billion base positions, and at each position there can be many possible variants. This represents a comprehensive annotation of the entire genome. You may be familiar with AlphaMissense and AlphaFold, which are also part of the broader annotation ecosystem, and we incorporated AlphaMissense into our database. To improve usability, we developed FAVOR-GPT, a large language model interface for FAVOR that leverages this massive relational database. This system relies on specialized SQL queries and efficient indexing, because it is not feasible to search for individual variants line by line using standard tools such as \texttt{R}. A sufficiently intelligent indexing strategy is essential.

The FAVOR annotator can be used to annotate any whole-genome sequencing dataset, including TOPMed, UK Biobank, and \emph{All of Us}. It is designed to be scalable and cost-efficient. For example, for UK Biobank, annotating over a billion variants costs less than \$25 as a one-time effort, and the annotations can then be reused for any phenotype or electronic health record outcome. We also developed the STAARpipeline, which integrates this annotation information directly into the analysis methods for whole-genome sequencing data in the UK Biobank. Analyzing any single phenotype again costs less than \$25. These efforts resulted in two papers published in \emph{Nature Genetics}. Together, they illustrate how critical annotation is. Annotation is not easy, but it represents the necessary last mile.

To summarize a few key points about data annotation: first, investment in data annotation requires courage. It took us five years to annotate these 9 billion variants, but such efforts can lead to transformative science, much like ImageNet did for AI. Second, it is essential to integrate domain-science-driven annotation into both statistical and machine learning method development. This integration promotes interpretability and produces results that are more useful in real-world applications, ultimately accelerating scientific discovery. Finally, scalability is critically important, both for annotation and for downstream analysis. Cloud computing is powerful but expensive, and we do not want researchers to break the bank. As a result, all the methods we develop must be scalable at very large cloud scale, with costs on the order of \$30 or \$50 to analyze a single phenotype in the UK Biobank. Scalability, therefore, is absolutely central to our work.

\paragraph*{Dan Nettleton:} Some obvious roles for statisticians involve developing methods for appropriately quantifying uncertainty, including uncertainty arising from limitations in the data acquisition process. Our discipline also has a strong track record of developing methods that provide interpretability and scientific understanding. Rather than simply focusing on predicting a response from multiple explanatory variables, statistical methods often yield insights into the relationships between explanatory variables and the response. Using more traditional statistical methods alongside empirical or algorithmic approaches can help achieve the best of both worlds. Furthermore, our expertise in experimental design and statistical process control can be leveraged to evaluate and monitor empirical or algorithmic methods and AI-driven decision-making pipelines. The current landscape also presents strong opportunities for collaboration with computer scientists, machine learning researchers, data engineers, and others beyond our usual collaborative partners.

\paragraph*{Abel Rodriguez:} Recently, I was reminded of a quote by George Box that goes something like, “Statisticians have the choice between being first-rate scientists or second-rate mathematicians.” To me, this quote resonates strongly with Bhramar’s and Xihong’s comments. Strong mathematical skills are an indispensable part of a statistician’s training, but blindly applying methods without understanding the context in which they are used is often a recipe for disaster. In my experience, the most impactful methodology is driven by challenges that arise in substantive, real-world problems. I have argued before that one of our strengths as statisticians is our careful attention to the data collection process and to sources of data bias, and I hope we can retain—and even strengthen—this as part of our core identity. For example, many issues grouped under the umbrella of fairness in AI and data science ultimately stem from a lack of attention to representativeness, causality, and confounding. In my department, we recently introduced an undergraduate course on the ethics of algorithmic decision-making in which these connections are explored in detail.

%I would also like to return to what Rebecca said about the need for changes in the promotion system. I absolutely agree with her comments and would go a step further: the broader culture needs to change.
The broader culture around our promotion systems needs to change. The message many of our students—especially Ph.D. students—receive is that the most “prestigious” careers are in academia, and academic hiring practices strongly favor method development. Sometimes this preference is stated explicitly, but more often it is implicit (for example, consider the types of candidates interviewed during your last departmental search). Similar dynamics arise in research funding: statisticians engaged in more applied work often fall through the cracks and struggle to secure funding because their work does not align neatly with existing disciplinary structures. Given this, it should not be surprising that many students are reluctant to engage in “data work.” Relatedly, many statistics departments have struggled to adapt to the reality that a large fraction of Ph.D. graduates now pursue diverse industry roles, with only a minority entering academia. Addressing this shift will require training students with stronger data skills, including a deep understanding of data curation, data wrangling, and data annotation.

\paragraph*{Eric Xing:} Returning to the long-standing debate between Kantians and Humeans—whether data (experience) or mind (reason) lies at the center of the definition of “knowledge”—this question has always been controversial. However, it is important to recognize that the balance can shift with the amount of data available or feasible to collect given technological advances, as well as with the power of the tools available for reasoning, such as natural laws and mathematical methods. The situation in Newton’s time and the present day are fundamentally different, marked by an unprecedented ability to obtain and generate data. In this context, adopting a data-centric attitude can be extremely helpful for rediscovering problems and redefining priorities. For example:

\begin{enumerate}
\item \textbf{On data labeling and “cleaning”:} Large language models are pre-trained on massive amounts of unlabeled data using a self-supervised learning mechanism known as next-token prediction. By defining the task in this way, all information is assumed to reside within the data itself, and all data—including “dirty” or “incorrect” data—are treated as part of the information content, provided the data collection process is sufficiently comprehensive and diligent. Through objectives such as masked language modeling implemented via autoregressive procedures, it is hypothesized that balanced, rich, and actionable abstractions—such as associations, dependencies, and redundancies—can be distilled into the model’s latent representations. These representations can then support a wide range of downstream generative and predictive tasks with high competence. In principle, no explicit supervision is required in this setting; supervision becomes necessary primarily in more advanced learning scenarios, such as reinforcement learning, that involve interaction and feedback. In classical statistical learning, labeling and cleaning may introduce additional bias imposed by the analyst, and it is not always clear that such practices add value when an alternative option is simply to collect more data. In this sense, focusing on the quality and breadth of data may be more valuable than focusing narrowly on the meaning of individual data points.

\item \textbf{On data collection and curation:} It is essential to remain aware of how data will be used and what tools will be employed to analyze it. Modern large-scale datasets differ substantially from the small datasets traditionally encountered in statistics: they are extremely high-dimensional (e.g., millions of dimensions in text data), multi-modal (combining continuous, discrete, categorical, and dynamic information across modalities), and massive in volume (often trillions of tokens). They undergo extensive technical transformations, including tokenization, normalization, alignment, and embedding, and frequently require domain-specific preprocessing. Moreover, in many modern AI systems, synthetic data are prevalent and often indispensable. These realities demand that individuals responsible for data collection, storage, indexing, curation, labeling, and processing possess a solid understanding of contemporary AI methodologies.

\item \textbf{On revisiting classical statistical concepts:} In the era of large data and large-scale computation, some conventional statistical concepts merit reconsideration. For example, why do we rely so heavily on the notion of a “distribution” and on addressing issues such as distributional shift? Why do we emphasize parametric models and sufficient statistics? These practices implicitly assume limited, relatively homogeneous data and constrained computational resources, necessitating abstraction as a means of storing knowledge. But what if we instead operate with extremely high-dimensional data, enormous sample sizes, and sufficient computational capacity to store, index, and manipulate data directly? In such settings, the data themselves may constitute the knowledge, and nonparametric approaches—such as $k$-nearest neighbors—may suffice for prediction via interpolation or extrapolation, either in the original data space or in a learned latent embedding space. In these regimes, tools such as manifold analysis and information-theoretic signal processing (e.g., compression, encoding, and decoding) may offer more actionable insights.
\end{enumerate}

\section{Engaging with Innovations in Empirical Modelling }

Statisticians should engage directly with modern machine learning and empirical modeling \citep{LinCaiDonohoEtAl2025StatisticsAI,Breiman2001TwoCultures,IMS2025PresidentialAddressStatisticsCrossroadsAI,he2025statistics}, rather than viewing them as external or purely engineering developments. The panel emphasized that statistical perspectives—uncertainty quantification, bias and causal thinking, interpretability, and rigorous evaluation—are essential for making these models robust, generalizable, and scientifically useful, especially in high-stakes settings. This will require greater fluency with contemporary ML practice and closer collaboration with computer scientists and data engineers.  

\paragraph*{Xihong Lin:} I will share some of my thoughts on this topic, and I will tell two stories. The first takeaway I would like to emphasize is that it is important for us to be open-minded, curious, willing to think outside the box, and ready to take risks. To illustrate this, I will tell a story about ChatGPT and large language models. As many of you know, ChatGPT-3 has about 175 billion parameters, and ChatGPT-4 has on the order of one trillion parameters. When ChatGPT first came out last December, my immediate reaction as a statistician was that it must be completely overfitting—the number of parameters is far larger than the sample size. How could this possibly work? But the mindset of engineers and AI researchers is quite different. Their approach is to make it work first and then figure out why. This reminds me of the Cox model. When Cox first proposed the partial likelihood, there was no formal theory. Based on his intuition, he believed the partial likelihood would work. About ten years later, the necessary theory was developed to provide a theoretical foundation for the Cox model. This is why I think we need to remain open-minded.

Later, I learned about an interesting phase transition phenomenon known as double descent. Statisticians are very familiar with the left-hand side of the curve. We know that as the number of parameters increases, the training error always decreases, which leads to overfitting. That is why we rely on test data. When the number of parameters increases to a certain point, prediction performance typically deteriorates due to overfitting. This is the classical regime. However, there is an intriguing double descent phenomenon: when models become extremely large, it is possible to find solutions within this vast model space for which the prediction error decreases again. This suggests that some of what we learned in graduate school may not always hold in modern settings. We need to step outside our comfort zone and be open to AI and engineering mindsets.

The second story comes from a recent experience at the JSM Young Researcher Workshop, where we had a lively discussion about statistics and AI. We asked ourselves how to think about truly disruptive statistics. Traditional statistical theory is being challenged by many emerging learning methods, such as transformers, large language models, and diffusion models. Over the past year, many of us have asked what statisticians can bring to the AI table. People often agree that statisticians can help build trustworthy AI and account for uncertainty and randomness. But then we must ask: how exactly do we do that? Are we interested in making inference on the trillions of parameters in large language models, which often lack clear interpretability? If not, how can we make inference on population parameters of interest that are not explicitly part of these models? How do we even define a population when models are trained on data aggregated from countless sources across the Internet? What do we mean by population parameters in this context, and how can they be connected to models with trillions of parameters?

Another topic we discussed concerns the very definition of data. This is a basic concept we all learned in statistics, yet text, video, and audio data from the Internet represent new forms of data that are not collected through traditional sampling schemes. Should synthetic data generated by models be considered data? These are deep and challenging questions for which we do not yet have clear answers. But they underscore the importance of rethinking and developing disruptive approaches to statistical inference.

\paragraph*{Eric Xing:} At the core of this question lie many fundamental tensions and disagreements between empiricism and rationalism, to which scientists—and statisticians in particular—consciously or unconsciously subscribe. This naturally raises questions such as: What is the purpose of machine learning research, and how should it be pursued? Framed in this way, it becomes easier to articulate how statisticians can be involved and contribute.

Even within the machine learning community, attitudes toward deep learning—the foundation of recent breakthroughs in large language models and generative AI—are mixed. Should the goal be to understand learning mechanisms (the \emph{why}) or to design and implement learning systems (the \emph{how})? Is statistical learning theory, which aims to provide mathematical guarantees, still useful—and if so, guarantees of what?

Empiricists question the intrinsic value of pure inductive reasoning, as it relies on assumptions such as continuity, uniformity, and stable associations in the real world—assumptions that cannot be proven within the same logical system. Many trailblazers in the machine learning community have chosen to bypass these philosophical debates and instead focus on building real systems at scale, emphasizing well-defined use cases and measurable utility. In this paradigm, abstract or philosophical definitions of reasoning are often replaced by functional or operational ones. For example, language reasoning based on next-token prediction—through substantial engineering effort—can support arbitrary context, content, and style, giving end users the experience of “intelligence” in certain interactions.

More recently, researchers have explored more advanced forms of operational reasoning, sometimes referred to as “reasoning via simulation.” From a lay perspective, this can be viewed as thought experiments or hypothetical reasoning. Such approaches rely on world models that simulate future states given current conditions and actions, enabling the generation of actions, multi-step plans, or strategies that optimize desired outcomes. These developments make extensive use of statistics and applied mathematics as enabling tools—for example, sampling via generative models, exploration via Monte Carlo tree search, content generation and concept extraction via diffusion models, and breaking long contexts into more tractable segments using state-space models. The role of statistics and statisticians in this ecosystem is clear. However, relevance and impact depend less on which tools one knows and more on the mindset governing how those tools are used and for what purposes.

Becoming overly consumed by endless debates about the ultimate objectives of empirical modeling can prevent statisticians from actively engaging in, understanding, and improving recent innovations in machine learning.

\section{Next-Generation Talent in Large-Scale Applications}

%\paragraph*{Questions: What are the best practices for training the next-generation statisticians for research in the field of large-scale data/AI applications?  Can you share examples of successful programs or initiatives that have helped nurture skills and expertise in this area?}

%\paragraph*{Xihong Lin:}

Training the next-generation statisticians must extend beyond traditional coursework to include end-to-end skills in data curation, modern ML, scalable computing, and principled inference about uncertainty, bias, and interpretability. The panel highlighted interdisciplinary curricula, team-based capstones with real datasets, cross-mentorship with CS and domain experts, and strong reproducible software practices as effective models \citep{LinCaiDonohoEtAl2025StatisticsAI,he2025statistics,Scharfmann2025PasteursQuadrant}. Just as importantly, departments should build supportive infrastructure and clear career pathways—often in partnership with industry and staff-scientist teams—to help trainees learn how to deliver trustworthy systems at scale.

\paragraph*{Bhramar Mukherjee:} From my personal viewpoint, we need to place greater emphasis on three C’s in our education: \emph{communication}, \emph{collaboration}, and \emph{computation}. Some data science and health data science programs have begun to address these areas at the master’s level, but we need much stronger emphasis on these skills in our graduate programs. In particular, we need dedicated Ph.D. tracks in health data science.

We also need to rethink the flavor of theory that we teach. Very few departments offer courses on the mathematical underpinnings of AI or on how to think about graphs and networks as stochastic objects. These topics are often treated as special topics courses, but they should instead become part of the expected core curriculum. Everyone should have at least an introductory course in deep learning and understand what transformers and autoencoders are.

Summer institutes and REU programs are exceptionally important. Many quantitatively inclined students are drawn to computer science, but if we can engage them early and intervene, many of them may be attracted to statistics and biostatistics.

Finally, it is critical to develop stronger ties with computer science and engineering. In my view, biostatistics departments in particular tend to have weaker connections with engineering, and this is something that is hurting us.

\paragraph*{Eric Xing:} We should not overload young students with what we believe to be important tools, but instead load them with problems and curiosity, and expose them to opportunities to work on real problems—clean or dirty. The focus should be on learning how to use and understand new tools, while also teaching interdisciplinary topics. Our current statistics curriculum has not changed substantially for decades; by contrast, curricula in computer science and artificial intelligence have undergone multiple major revisions and continue to evolve rapidly. It is therefore necessary to assess how much of what we teach is actually used in practice, and how many subjects encountered in graduates’ professional lives are not covered in the classroom, and then make adjustments accordingly. Sitting in offices with the same group of scholars for decades, debating—for example—how to teach a one-year regression course using a different textbook, is unlikely to generate truly new ideas. It is time to think seriously about what should be removed from the current curriculum before deciding what to add.

As an example, in a recent effort to create a new undergraduate program in AI, we envisioned that students would not only develop deep technical expertise in areas such as data science, machine learning, natural language processing, computer vision, and robotics, but would also receive cross-disciplinary training in industrial design, market analysis, consumer studies, management, finance, and communication. In addition, the program emphasizes a hands-on, experiential curriculum designed to cultivate an entrepreneurial mindset, preparing students to lead AI-driven transformations in society and industry. The goal is to train AI-native students who can both master complex technical challenges and emerge as visionary, multifaceted problem solvers. A similar level of reinvention is also necessary for statistics programs.

\section{Engaging AI Stakeholders}

Statistics should engage AI stakeholders more proactively by communicating its distinctive value—uncertainty quantification, bias and causal reasoning, interpretability, and reproducible evaluation—in AI-relevant terms. The panel emphasized building a robust ecosystem through sustained cross-sector partnerships, shared data and benchmarking infrastructure, and training pathways that connect academia and industry \citep{WhiteHouse2025GenesisMissionFactSheet}. They also stressed the need to advocate for funding and review structures that reward end-to-end contributions, enabling statistical principles to shape trustworthy, real-world AI.

\paragraph*{Bhramar Mukherjee:} Institutional leadership in data science is critically important for statisticians to pursue. In my role as Associate Vice President for Research Data Strategy, I participated last year in Michigan’s investments in AI, high-performance computing, and research data strategy. Simply listening to conversations across the university provides a valuable big-picture perspective, and these roles also offer direct access to senior leadership, including the Vice President for Research, the Provost, and the President. Investments made at this level are often far larger than what individual schools or departments can typically envision.

It is also important to build partnerships with national laboratories that operate large-scale supercomputing resources.

\noindent\textbf{Ten reminders of what statisticians bring to the table:}
\begin{itemize}
\item Think carefully about bias in training data and transportability to test or deployment data.
\item Consider sampling strategies and experimental design principles for AI systems.
\item Emphasize uncertainty quantification.
\item Focus on rigorous evaluation and comparison of AI tools and models.
\item Accept that statisticians are unlikely to build better large language models themselves.
\item Accept that many problems are theoretically intractable, and make peace with this intrinsic discomfort.
\item Collaborate rather than merely criticize, compete, or feel insecure.
\item Recruit more faculty and staff with engineering backgrounds into statistics and biostatistics departments.
\item Forge strong partnerships with national laboratories and industry.
\item Use AI to achieve positive global impact by leveraging domain expertise.
\end{itemize}

\paragraph*{Dan Nettleton:} Over the years there are many examples of statisticians who have delved deeply into understanding real problems from start to finish, even though many aspects of such work go beyond our traditional roles.  To continue to be relevant and valued, we need to get to know the real problems from start to finish, get to know the people working on the real problems, ask questions, get involved in developing solutions, and tell our stories.

\paragraph*{Eric Xing:} First of all, put yourself in the shoes of those from whom you are seeking funding or promotion. Why should they invest, say, \$10 million in you? Why should they prioritize your project over others? What returns—financial or intellectual—do they receive? What impact can they clearly see? Many may argue that utility is not everything in academia, but for the sake of the community, a proper balance between idealism and pragmatism, and between entitlement and responsibility, is essential.

More specifically, here are a few points we can advocate to address the question at hand:
\begin{enumerate}
\item Continue to solve problems with real impact, and position ourselves at the forefront as solution providers and doers, rather than becoming overly focused on showcasing clever intellectual acrobatics or repeatedly analyzing others’ operational results without producing our own.
\item Do not be overly selective about problems. Earn the freedom to pursue any research topic by demonstrating value, rather than insisting on entitlement to that freedom. Show real utility and impact through your work—keep your head down, get your hands dirty, step out of the office and into the field. After all, this is where statistics was born.
\item If your true passion lies in pure theory, then develop theory that explains big, consequential phenomena—for example, uncertainty quantification for a given large language model, or learning curves that characterize the sample and computational complexity of real-world AI systems in operation.
\end{enumerate}

\section{Discussion by David Donoho}
\begin{figure}[htbp]
    \centering
    \includegraphics[width=0.9\textwidth]{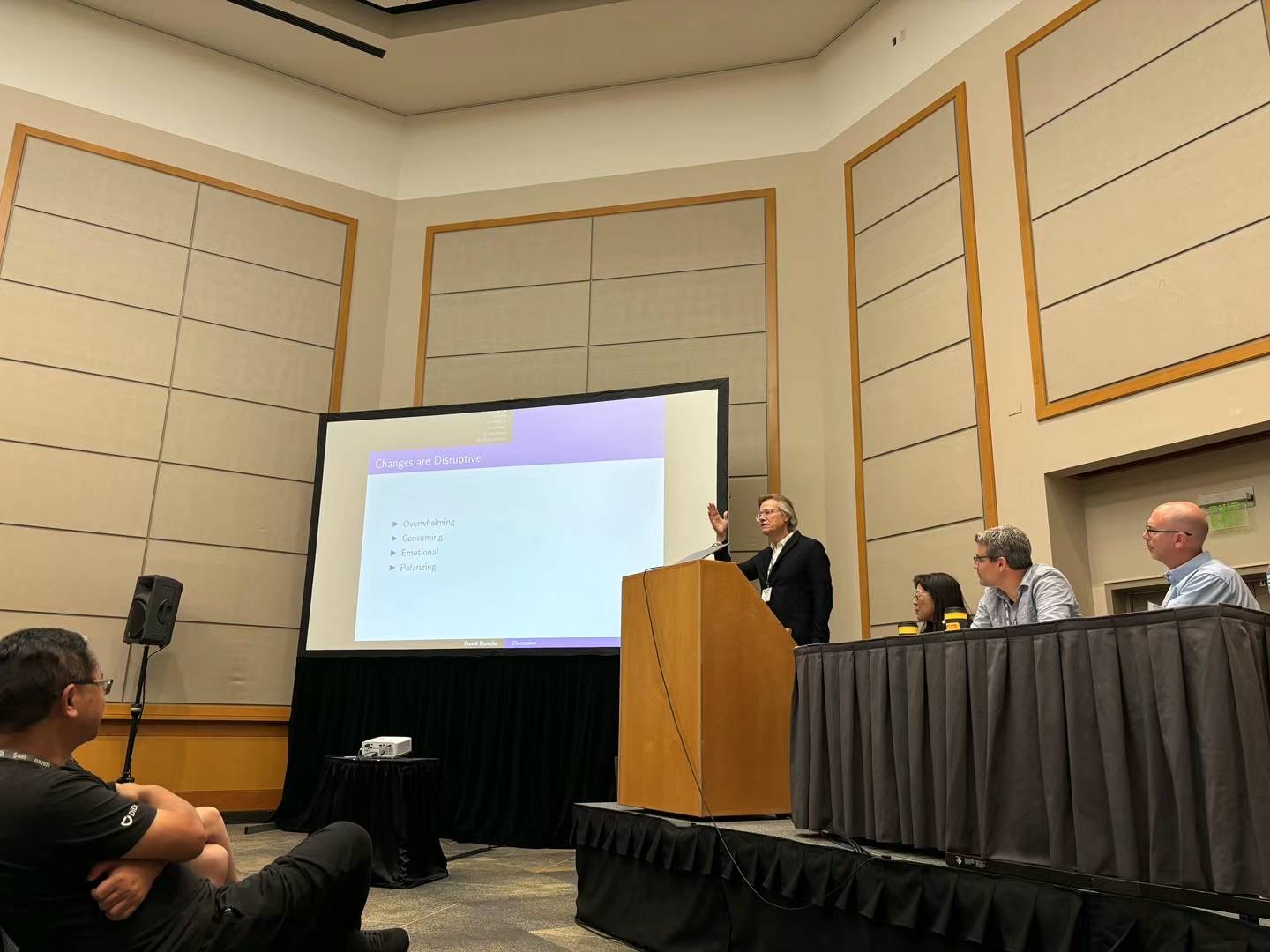}
    \caption{Discussion by David Donoho in the 2024 JSM town hall
    ``Statistics in the Age of AI''.
    The discussion brought together perspectives from statistics, biostatistics,
    and machine learning.}
    \label{fig:townhall_panel}
\end{figure}

\paragraph*{David Donoho:} I would like to make a few comments. Eric comes from a different community, and it is very important for us to hear what he has said. Practically every point he made was solid gold. It may be that not everyone here immediately sees it that way, but he was, in my view, checking every box where statistians are currently in denial. He articulated, statement after statement, things that statisticians ought to be feeling in their bones but often are not. I wanted to call that out explicitly and give him a strong shout-out and thank you.

I am not exactly sure what is most up to date at the moment, but many of the other points raised were extremely sharp and to the point. It is almost impossible for me to cluster them coherently, so I will just be brief and touch on a few background ideas.

We all know that the Olympics are going on right now. At the Olympics, you see crowds, and you see people taking pictures of those crowds. There is a picture circulating of an older gentleman—Mick Jagger—surrounded by much younger people, all of whom are looking at their phones. Mick Jagger might as well not exist for them; whatever is happening on their phones is far more important. Meanwhile, Mick Jagger himself is looking out at the action—perhaps women’s gymnastics or something similar—while everyone around him is completely absorbed in their devices. A picture like this could not have been taken 20 years ago. The media we use and the ways we obtain information have changed human behavior and human minds, and the extent of this change is crushing. We do not yet fully understand it.

Some of Eric’s comments were getting at this point. These are not the young people of 20 years ago. The minds they have are not the minds we had. We were never them—most of us, at least—and perhaps there are some very young people here, and I apologize if I miss you. But the world has changed, and it has changed comprehensively. There are many ways to think about this, and we are only beginning to see it clearly. These changes are extremely disruptive. I experience them as overwhelming, utterly consuming of attention, highly emotional, and highly vulgarizing.

Many statistics departments have rebranded themselves as “statistics and data science,” as if these two things can simply be bolted together. That is not going to work, precisely because of the differences involved. If we go back to the earlier image, Mick Jagger is like R. A. Fisher. They are both legends; the only difference is that one is in the Rock \& Roll Hall of Fame and the other is a Fellow of the Royal Society. But just as the people in that crowd could not be induced to notice Mick Jagger, they also would not be able to engage with or understand R. A. Fisher. Let that sink in for a moment.

This connects to ideas discussed in works such as those on the “Gutenberg parenthesis.” Literacy as we know it emerged with the mass production of books and may be ending with the widespread adoption of smartphones. We were even asked earlier whether it is necessary to develop traditional forms of literacy in an era when large language models are readily available. I do not know what will be required for high-quality thinking in the future. I only know what shaped my own mind: books, reading, and sustained engagement with text. My personal opinion does not really matter, because this mode of engagement may simply not exist for future generations. Information technology delivered directly through devices—perhaps even through glasses—will be overwhelmingly dominant.

I wanted to bring this up to encourage reflection on the wisdom shared today. Many of the points raised are fundamental and deeply important. At the same time, it is hard to act on them. Many of us know what we ought to do, but it is difficult to mobilize the urgency and energy required to actually do it. Yet we must. One key insight from the idea of the Gutenberg parenthesis is that mathematics itself is part of that parenthesis. Statistics is a fundamentally mathematical discipline, and it, too, sits within the Gutenberg parenthesis. We are now beyond its closing. For most people on Earth, it has already closed. Smartphones have reached billions of people in the last decade who will never experience extended engagement with print culture or with the kind of education many of us grew up with.

That is the urgency I feel. There were so many insightful comments today that I cannot possibly do them all justice. But thank you all for what you shared. I have benefited enormously from this discussion, and I would like to put a metaphorical “heart” on every single one of the panelists.

\spacingset{1.05}
\bibliography{references}

\end{document}